\documentclass[lettersize,journal]{IEEEtran}
\usepackage{amsmath,amsfonts}
\pdfoutput=1
\usepackage{array}
\usepackage[caption=false,font=normalsize,labelfont=sf,textfont=sf]{subfig}
\usepackage{textcomp}
\usepackage{stfloats}
\usepackage{url}
\usepackage{verbatim}
\usepackage{graphicx}
\usepackage{cite}
\usepackage{multirow}
\usepackage{booktabs}
\usepackage{bbding}
\usepackage{pifont}
\usepackage{listings}
\usepackage{threeparttable}
\usepackage{makecell}

\begin{document}

\title{ChatTraffic: Text-to-Traffic Generation via Diffusion Model}

\author{Chengyang Zhang, Yong Zhang\thanks{Yong Zhang is the corresponding author.}, Qitan Shao, Bo Li, Yisheng Lv, Xinglin Piao, and Baocai Yin
        % <-this % stops a space

\thanks{Chengyang Zhang, Yong Zhang, Bo Li; Xinglin Piao; Baocai Yin are with Beijing Key Laboratory of Multimedia and Intelligent Software Technology, Beijing Artificial Intelligence Institute, Faculty of Information Technology, Beijing University of Technology, Beijing, China, 100124. (Cy\_Zhang@emails.bjut.edu.cn; zhangyong2010@bjut.edu.cn; shaoqt@emails.bjut.edu.cn; bo\_li@emails.bjut.edu.cn; piaoxl@bjut.edu.cn; ybc@bjut.edu.cn)}% <-this % stops a space
\thanks{Yisheng Lv is with the Institute of Automation, Chinese Academy of Sciences. (yisheng.lv@ia.ac.cn)}
\thanks{The research project is partially supported by the National Key R\&D Program of China (No. 2021ZD0111902), National Natural Science Foundation of China (No.62072015, U21B2038, U19B2039, 61902053) and Beijing Natural Science Foundation (4222021).}
}

\markboth{Journal of \LaTeX\ Class Files,~Vol.~14, No.~8, August~2021}%
{Shell \MakeLowercase{\textit{et al.}}: A Sample Article Using IEEEtran.cls for IEEE Journals}


\maketitle

\begin{abstract}
Traffic prediction is one of the most significant foundations in Intelligent Transportation Systems (ITS). Traditional traffic prediction methods rely only on historical traffic data to predict traffic trends and face two main challenges. 1) insensitivity to unusual events. 2) limited performance in long-term prediction. In this work, we explore how generative models combined with text describing the traffic system can be applied for traffic generation, and name the task Text-to-Traffic Generation (TTG). The key challenge of the TTG task is how to associate text with the spatial structure of the road network and traffic data for generating traffic situations. To this end, we propose ChatTraffic, the first diffusion model for text-to-traffic generation. To guarantee the consistency between synthetic and real data, we augment a diffusion model with the Graph Convolutional Network (GCN) to extract spatial correlations of traffic data. In addition, we construct a large dataset containing text-traffic pairs for the TTG task. We benchmarked our model qualitatively and quantitatively on the released dataset. The experimental results indicate that ChatTraffic can generate realistic traffic situations from the text. Our code and dataset are available at \url{https://github.com/ChyaZhang/ChatTraffic}.
\end{abstract}


\section{Introduction}
\IEEEPARstart{T}{raffic} prediction is a fundamental and pivotal task within the realm of Intelligent Transportation Systems (ITS). Its primary objective is to predict future traffic situations based on historical data \cite{1,2,3}. This task plays a significant role in facilitating peak flow warnings, alleviating congestion, and optimizing travel routes. Urban transportation systems, due to their complexity, are susceptible to various influencing factors, including weather, traffic accidents, road construction, etc. Consequently, the consideration of these diverse factors is of paramount importance in enhancing the accuracy of traffic prediction.

\begin{figure}[t]
  \centering
   \includegraphics[width=1.\linewidth]{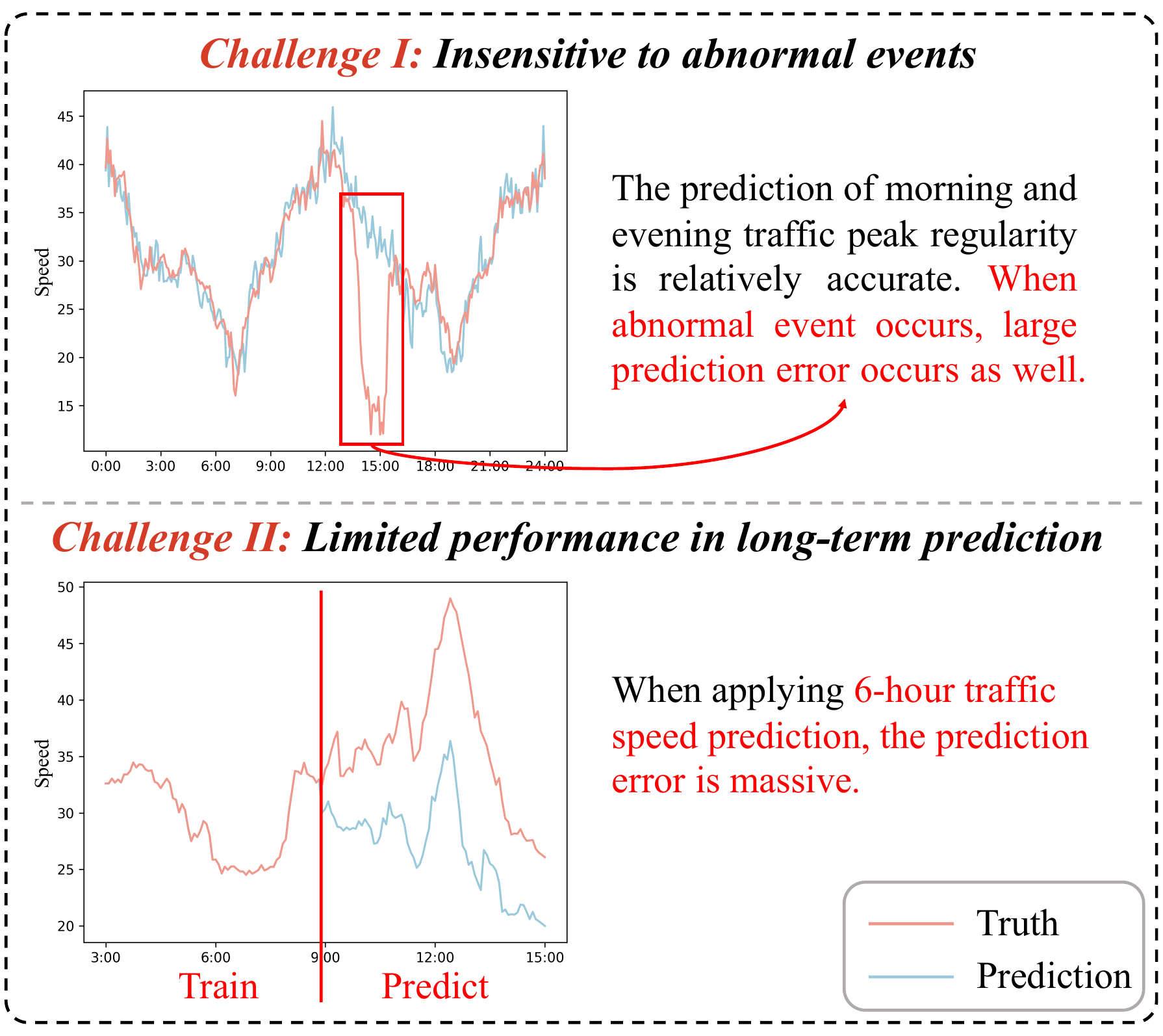}
   \caption{Two main challenges confronted by existing traffic prediction methods. 1) Insensitive to abnormal events. 2) Limited performance in long-term prediction.}
   \label{fig2}
\end{figure}

Most traffic prediction works use historical data to predict future data \cite{1,2,3,16,17,21,22}. Although these methods have demonstrated advanced short-term prediction capabilities on specific datasets, they still confront two main challenges, as depicted in Figure \ref{fig2}. 1) Insensitive to abnormal events. Real-world urban transportation systems frequently experience abnormal events, such as car accidents, road construction, and extreme weather, which can significantly disrupt traffic situations. Consequently, these events lead to deviations from the typical traffic patterns. Methods trained solely on historical data often struggle to provide accurate predictions when such abnormal events occur. 2) Limited performance in long-term prediction. Long-term prediction plays a significant role in traffic management. While existing traffic prediction methods excel in short-term predictions, typically within 30 minutes, there remains considerable room for enhancement in long-term prediction. Therefore, existing traffic prediction methods are often unable to address many practical scenarios. For instance, a concert will be held at the Beijing Workers' Stadium next Saturday night. The traffic situation near the stadium may be much different from the normal patterns. Traditional prediction methods are not suitable for anticipating traffic situations for this specific Saturday evening scenario.

\begin{figure*}[t]
  \centering
   \includegraphics[width=\textwidth]{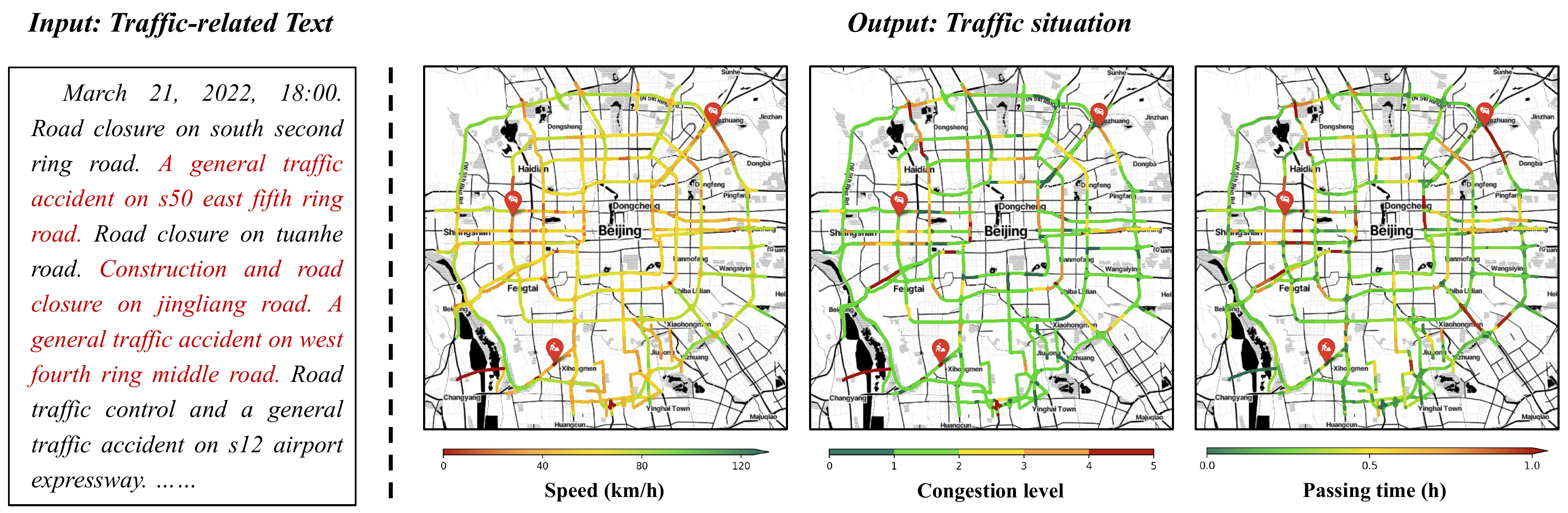}
   \caption{Text-to-traffic generation via diffusion model. Given a piece of text describing the transportation system (including time and events), we present the diffusion-based ChatTraffic to generate the traffic situation. For the first time, our proposed ChatTraffic is capable of generating traffic situations (speed, congestion level, and passing time) according to the text. This enables ChatTraffic to provide predictions of how future events (road construction, unexpected accidents, unusual weather) will affect the urban transportation system, pushing this domain a considerable step forward.}
   \label{fig1}
\end{figure*}

To address the limitations of traditional traffic prediction methods, some researchers have already integrated traffic data with text for traffic generation in previous studies. For instance, Huo \textit{et al.} \cite{4} construct a dataset containing the traffic-related text data collected from social media and the corresponding passenger flow data, and propose T$^{2}$GAN generating traffic situations from the text. However, T$^{2}$GAN necessitates the division of the city into regular grids for traffic generation, which significantly constrains its applicability. Additionally, these texts are not specific enough to describe the traffic situations, such as  ``The overall traffic situation in Beijing is good. The passenger flow is small.". Moreover, T$^{2}$GAN requires both text and traffic data as inputs in the inference phase. To this end, we explore a novel multimodal traffic prediction task called Text-to-Traffic Generation (TTG), as shown in Figure \ref{fig1}. In TTG, we not only leverage traffic data as input for training but also incorporate textual descriptions depicting the concurrent traffic situations, while only text is needed for inferencing. To better promote multimodal learning in the field of traffic prediction, we also construct a substantial traffic dataset containing detailed text descriptions for the TTG task.

To associate traffic-related text with traffic data, generative models provide a feasible way. Among them, Generative Adversarial Networks (GAN) based methods \cite{5,6,7} implicitly fit the data distribution via adversarial training to achieve high-quality generation. However, GAN-based methods are always restricted by mode collapse and unstable training. On the other hand, Variational Autoencoder (VAE) based methods \cite{8,9} are trained relatively stable but suffer from blurred details and offer low-quality outputs compared to GAN. Compared to GAN and VAE, diffusion models exhibit greater ease of training and superior generative capabilities, firmly establishing themselves as one of the most robust generative models available today. Diffusion-based methods \cite{10,11,12,13,14} have been proven successful in numerous generative tasks, especially in generating images from text. Inspired by the success of diffusion models, we believe that diffusion models also have the potential to address the TTG task well.

In this paper, we frame the traffic generation as a series of diffusion steps and introduce ChatTraffic, a simple yet effective framework based on Latent Diffusion Model (LDM) \cite{13}, for the TTG task. To overcome the challenges confronted by traditional traffic prediction methods, texts containing time and events are applied to guide the denoising process to achieve traffic generation. Furthermore, we augment a diffusion model with the Graph Convolutional Network (GCN) \cite{15}, a commonly employed network in traffic prediction methods. Except for time and events, traffic situations are also affected by the structure of the road network. In view of this, the primary idea of introducing GCN is to utilize the spatial information of the road network as a constraint to adjust the traffic features for more accurate conditional traffic generation. As shown in Figure \ref{fig1}, the proposed ChatTraffic is capable of generating traffic situations (speed, congestion level, and passing time) from the text. 

\begin{itemize}
\item[$\bullet$] We explore a novel traffic prediction task called Text-to-Traffic Generation (TTG). In this task, the first diffusion-based text-to-traffic generation model is proposed, named ChatTraffic.

\item[$\bullet$] We augment the diffusion model with the Graph Convolutional Networks (GCN), which leverage the spatial information inherent in the road network, leading to more realistic and accurate traffic generation.

\item[$\bullet$] We construct the first substantial text-traffic dataset for the TTG task. The dataset covers 1,260 roads within the fifth ring road area of Beijing, providing more than 20,000 text-traffic pairs.
\end{itemize}

The rest of the paper is structured as follows. Section \ref{section2} delves into the discussion of related works, while section \ref{section3} provides a comprehensive overview of both the diffusion model and the proposed ChatTraffic. Section \ref{section4} presents the experimental results along with a detailed analysis. Lastly, Section \ref{section5} encompasses the summary and outlook.

\section{Related Works}
\label{section2}
In this section, we offer a brief overview of the current research status in traffic prediction and relevant studies on social media data-related traffic applications. Additionally, the recent researches on text-based generative models are also reviewed.

\subsection{Traffic Prediction}
Traffic prediction can assist governments in better managing the urban transportation system and holds substantial importance within contemporary intelligent transportation systems. Traffic prediction methods are broadly classified into two categories: machine learning methods and deep learning methods.  In the early stages of research, machine learning theories are predominantly employed to formulate traffic prediction models \cite{16,17,18,19,20}. An early attempt \cite{16} proposes the seasonal Auto-Regressive Integrated Moving Average (ARIMA) to analyze the properties of highway traffic flow. Jeong \textit{et al.} \cite{19} introduce the Online Learning Weighted Support-Vector Regression (OLWSVR) to consider the time difference between traffic flow data. 

With the continuous development of urban transportation systems, these methods can no longer cope with the massive and complex traffic data. To this end, deep learning methods \cite{1,2,3,21,22,23,24,25,26} have gradually become the mainstream of research. Typically, Long Short-Term Memory Networks (LSTM) \cite{21,22} are applied to extract temporal correlations in traffic prediction. Shi \textit{et al.} \cite{25} propose a convolutional LSTM to effectively capture spatio-temporal information in traffic data. Fu \textit{et al.} \cite{24} associate the LSTM and Gated Recurrent Unit (GRU) to predict the short-term traffic flow. To better represent the irregular traffic network structure, the researchers integrate GCN into the traffic prediction task to extract spatial correlations. Specifically, the Spatio-Temporal Graph Convolutional Networks (STGCN) \cite{1} pioneers the combination of graph convolution and gated causal convolution to tackle the time series prediction problem in the traffic domain. Wu \textit{et al.} \cite{2} introduce Graph WaveNet (GWN), an innovative graph neural network designed for modeling spatial-temporal graphs. This model innovatively employs an adaptive dependency matrix, learned via node embedding, enabling precise identification of hidden spatial dependencies within the data.

Current traffic prediction methods have achieved good short-term prediction performance, but long-term prediction remains an unsolved challenge. Despite the intricate nature of transportation systems, most traffic prediction methods rely solely on a single data type, neglecting the potential benefits of heterogeneous data. In this paper, we aim to empower the model with better long-term prediction performance and the ability to perceive abnormal events by leveraging multimodal data.

\subsection{Social Media Data-related Traffic Applications}
In recent years, some researchers have expanded the scope of traffic prediction beyond just historical traffic data, incorporating additional sources such as social media data. These approaches have proven to be beneficial in enhancing the study and application of intelligent transportation systems. Typically, Ni \textit{et al.} \cite{51} develop a systematic methodology for scrutinizing social media activities and detecting event incidents. Chen \textit{et al.} \cite{52} utilize the continuous bag-of-words model to acquire word embedding representations from a vast dataset consisting of three billion microblogs, and propose LSTM-CNN to extract microblogs relevant to traffic, using the acquired word embeddings as inputs. Moreover, Yao \textit{et al.} \cite{53} suggest the utilization of Twitter messages as a probing technique for comprehending the influence of individuals' work and rest patterns during the evening and midnight from the previous day to the following morning on traffic conditions. Wang \textit{et al.} \cite{54} propose a method to alleviate the issue of data sparsity by integrating traffic event signals extracted from social media with GPS probe data. 

The above studies demonstrate the effectiveness of social media data in assisting traffic prediction and identifying abnormal traffic events. However, most of these studies concentrate on the statistical analysis of social media content, largely overlooking the rich semantic information presented on social media platforms. Therefore, the key to enhancing traffic situation generation lies in effectively correlating text data related to traffic situations with actual traffic data.

\subsection{Text-based Generative Models}
In the realm of computer vision, the task of generating images from text has been extensively explored. In the earlier stages, methods based on GAN \cite{5,6,7,27,28,29,30} demonstrate strong capabilities in associating text descriptions with images. Reed \textit{et al.} \cite{27} develop a novel architecture and GAN formulation for visual concept translation from characters to pixels. Stack-GAN \cite{28} generates realistic images based on text descriptions. Xue \textit{et al.} \cite{29} propose the Attention Generative Adversarial Network (AttnGAN), which incorporates an attention mechanism into GAN to generate fine-grained image details. Zhu \textit{et al.} \cite{30} design a memory write gate to selectively filter crucial segments of text and introduce the Dynamic Memory Generative Adversarial Network (DM-GAN) for generating high-quality images using textual descriptions.

Diffusion models \cite{10,11} are emerging generative models capable of synthesizing high-quality images. Following the pioneering work of leveraging diffusion steps for data distribution learning \cite{41}, diffusion models have demonstrated remarkable capabilities in various synthesis tasks, such as image editing \cite{33,34,35}, image in-painting \cite{14,37,38}, image super-resolution \cite{39,40}, text-to-image generation \cite{13,31,32,36}, and other applications \cite{46,47,48,49}. Our work is mainly related to text-guided synthesis tasks. In the text-to-image generation task, the diffusion-based methods demonstrate performance beyond that of GAN. Typically, Saharia \textit{et al.} \cite{32} propose Imagen, which achieves profound language understanding with photo-realistic image generation. Nichol \textit{et al.} \cite{14} present Guided Language to Image Diffusion for Generation and Editing (GLIDE), a diffusion model that leverages textual guidance for both realistic image generation and image editing. 

These text-to-image generation models provide a potential idea for generating traffic situations. Inspired by the text-to-image diffusion model, we propose the first diffusion-based method for text-to-traffic generation by associating traffic data with text. Text-to-traffic generation can improve the robustness of prediction results to abnormal events, making it more suitable for real-world applications.

\section{Methodology}
\label{section3}
In this section, we first briefly summarize how the TTG task differs from the traditional traffic prediction task in section \ref{section3.1}. We then introduce the diffusion model for traffic generation in section \ref{section3.2}. Finally, we illustrate the proposed ChatTraffic in section \ref{section3.3}. Given a piece of text describing the traffic system (including time, location, and events), the goal of ChatTraffic is to generate traffic situations (speed, congestion level, and passing time) that align well with the text.

\begin{figure*}[t]
  \centering
   \includegraphics[width=\textwidth]{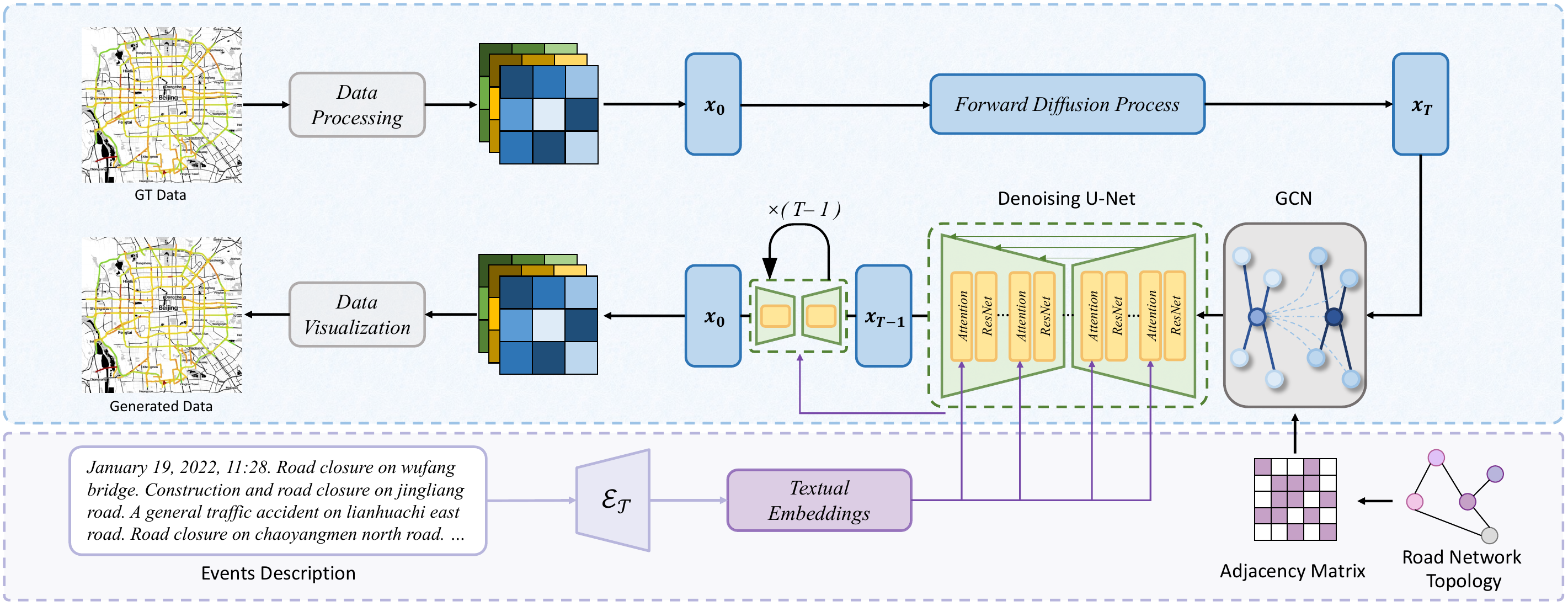}
   \caption{Method overview. The core components of our proposed ChatTraffic are a UNet consisting of ResNet and cross-attention, and a GCN. We first populate and reshape the data to make it more suitable for use as input to a diffusion model. We use a text encoder to extract feature embeddings from text describing the traffic system. Furthermore, we introduce a GCN to achieve stronger generative consistency. The GCN takes the noisy traffic data $x_{t}$ and the adjacency matrix $A$ describing the spatial correlations of the road network as inputs to associate structural and state features of the road network. The UNet takes the textual feature embeddings and the outputs of the GCN as inputs to predict the denoised traffic data.}
   \label{fig3}
\end{figure*}

\subsection{Motivation and Task Definition}
\label{section3.1}
Traditional traffic prediction methods rely on historical data and lack the integration of multimodal features. Consequently, these methods are limited to forecasting regular traffic patterns, such as morning and evening peaks. To enhance the robustness of the prediction model to abnormal events, we explore a multimodal traffic prediction task called Text-to-Traffic Generation (TTG) and introduce a feasible solution for it. 

The key challenge of the TTG task is how to associate text with the spatial structure of the road network and traffic data for generating traffic situations. For example, given text describing “A general traffic accident on east third ring middle road.”, the generated traffic situation is supposed to reflect slower speeds on the east third ring middle road. In view of this, users, especially traffic managers, who are concerned with the traffic situations of a certain scenario or event, can simply provide a text description without any other specialized and complex operations. In this work, we construct a substantial text-traffic dataset and propose a diffusion-based framework coupled with GCN to address this challenge. 

\subsection{Traffic Generation Diffusion}
\label{section3.2}
We present an overview of ChatTraffic in Figure \ref{fig3}, which is built upon the diffusion model. Diffusion models \cite{10,11,41} are generative models capable of generating high-quality content. Its basic form consists of two Markov chains in opposite directions while two processes are transiting through the chain. The forward diffusion process gradually adds noise to the traffic data until it is completely corrupted by Gaussian noise and becomes indistinguishable. The inverse process focuses on learning how to recover the original data distribution from this noise. 

\subsubsection{Forward diffusion process}
Given the clean traffic data $x_{0}$ sampled from a real data distribution $q(x)$, we gradually add a total of $T$ steps of Gaussian noise to $x_{0}$, obtaining a series of variables $x_{1}, x_{2}, ..., x_{T}$. In the forward process, the data $x_{t}$ is only related to the previous moment data $x_{t-1}$ and can be expressed as
\begin{equation}
  q\left(x_{t} \mid x_{t-1}\right)=\mathcal{N}\left(x_{t} ; \sqrt{1-\beta_{t}} x_{t-1}, \beta_{t} I\right),
\end{equation}
where $\beta_{t}$ determines the mean and variance of the added noise and satisfies $\beta_{1} \textless \beta_{2} \textless ... \textless \beta_{T}$. This indicates that as $t$ increases, the added noise progressively grows larger. Since it is defined as a Markov chain, the joint distribution of $x_{1:T}$ for a given $x_{0}$ is
\begin{equation}
    q\left(x_{1: T} \mid x_{0}\right)=\prod_{t=1}^{T} q\left(x_{t} \mid x_{t-1}\right).
\end{equation}

To avoid iterative computation, we can directly compute $x_{t}$ from $x_{0}$ by utilizing the reparameterization technique. Given $\alpha_{t} = 1- \beta_{t}$ and $\bar{\alpha}_{t}=\prod_{i=1}^{t} \alpha_{i}$ , $x_{t}$ can be represented as
\begin{equation}
    x_{t}=\sqrt{\bar{\alpha}} x_{0}+\sqrt{1-\bar{\alpha}_{t}} \epsilon_{t},
\end{equation}
where $\epsilon_{t}$ is the noise used to destroy $x_{t}$ and $\epsilon_{t} \sim \mathcal{N}(0, I)$. The relationship between $x_{t}$ and $x_{0}$ is denoted as
\begin{equation}
     q\left(x_{t} \mid x_{0}\right)=\mathcal{N}\left(x_{t} ; \sqrt{\bar{\alpha}} x_{0},\left(1-\bar{\alpha}_{t}\right) I\right).
\end{equation}

\subsubsection{Inverse generative process}
The inverse process recovers the original traffic data from the Gaussian noise and is also a Markov chain process. Since the noise we add each time in the forward process is small, we can use a Gaussian transform $p_{\theta}\left(x_{t-1} \mid x_{t}\right)$ parameterized by the neural network $\theta$ to recover $x_{t-1}$ from $x_{t}$, which is represented as
\begin{equation}
    p_{\theta}\left(x_{t-1} \mid x_{t}\right)=\mathcal{N}\left(x_{t-1} ; \mu_{\theta}\left(x_{t}, t\right), \sum_{\theta}\left(x_{t}, t\right)\right),
\end{equation}
where $\mu_{\theta}\left(x_{t}, t\right)$ and $\sum_{\theta}\left(x_{t}, t\right)$ are the predicted mean and covariance at time step $t$. To achieve higher synthesis quality, Ho \textit{et al.} \cite{42} estimate $\epsilon_{\theta}\left(x_{t}, t\right)$ rather than directly predict $\mu_{\theta}\left(x_{t}, t\right)$. Based on Bayesian theory, $\mu_{\theta}\left(x_{t}, t\right)$ can be expressed as
\begin{equation}
    \mu_{\theta}\left(x_{t}, t\right)=\frac{1}{\sqrt{\alpha_{t}}}\left(x_{t}-\frac{\beta_{t}}{\sqrt{1-\bar{\alpha}_{t}}} \epsilon_{\theta}\left(x_{t}, t\right)\right).
\end{equation}

\subsection{ChatTraffic}
\label{section3.3}
Given a text $y$ describing the traffic system, ChatTraffic denoise the data $x_{t}$ to the final noise-free $x_{0}$. We first process the data to make it more suitable as the input to the diffusion model. Concurrently, the provided text is encoded into feature embeddings using a text encoder. Combined with the cross-attention mechanism, GCN and Unet, ChatTraffic predicts $\epsilon_{\theta}\left(x_{t}, t\right)$ to get the cleaner $x_{t-1}$. After $T$ time steps, the noise-free $x_{0}$ is obtained. With data visualization, we present the traffic situations on the map to provide a more intuitive representation.

\subsubsection{Data processing}
\begin{figure}[t]
  \centering
   \includegraphics[width=1.\linewidth]{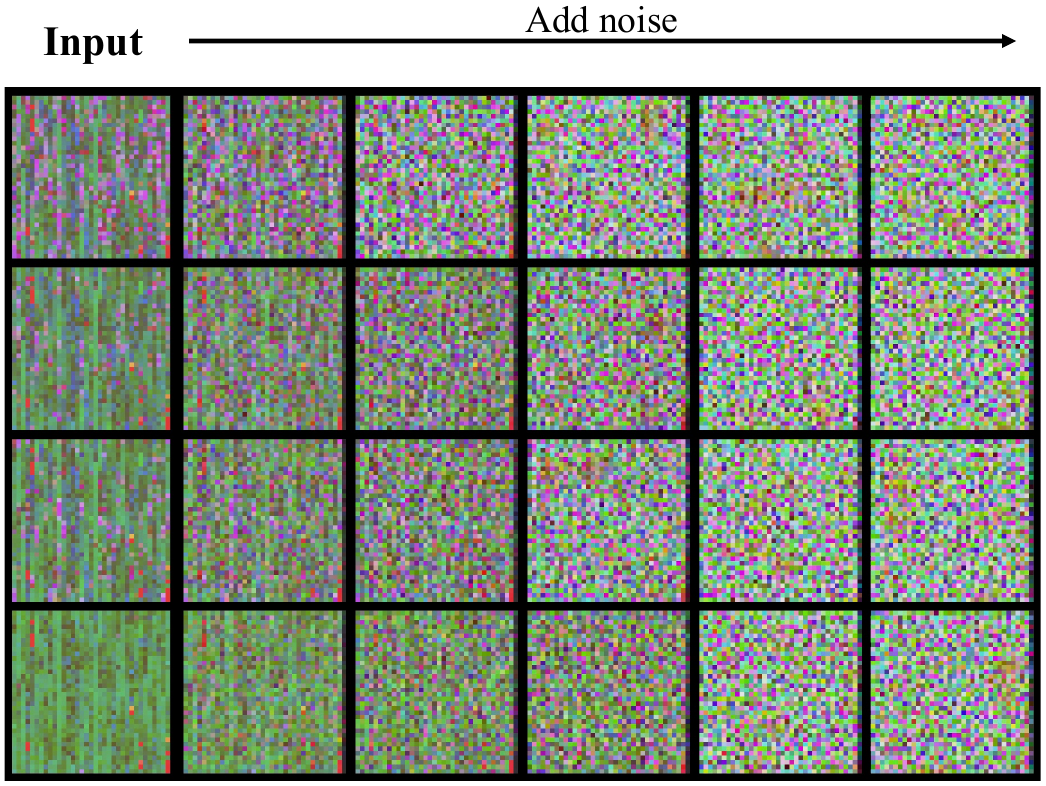}
   \caption{Illustration of adding noise to traffic data. From left to right represents the gradual addition of noise to the traffic data. From top to bottom are four different junctures of traffic data presented in the form of `images'.}
   \label{fig4}
\end{figure}

The released dataset comprises over 20,000 text-traffic pairs within the fifth ring road in Beijing. Each data $x$ contains three types of features (speed, congestion level, passing time) derived from 1260 roads, where $x \in \mathbb{R}^{N \times d}$, $N=1260$, and $d=3$. However, there exists a considerable disparity in the dimensions of $x$, rendering it unsuitable as input for the diffusion model. To address this issue, we rearrange the traffic data $x$ in the form of images. Specifically, we define 36 additional ``empty roads" and pad them into $x$ to make $x \in \mathbb{R}^{N' \times d}$, $N'=1296$. $x$ is then reshaped to $x \in \mathbb{R}^{H \times W \times d}$, where $H$ and $W$ equal to 36. If we view $x$ as an image, the three channels represent speed, congestion level, and passing time, and each pixel represents a road. In Figure \ref{fig4}, we illustrate how the traffic data $x$ evolves throughout the forward process of ChatTraffic.

\subsubsection{U-Net \& textual encoder}
To associate text with the traffic data, we formulate the TTG task as a conditional generation problem and implement it using a modified LDM. LDM transforms the data $x$ into the latent space to obtain $z$, achieved through an encoder $\mathcal{E}(\cdot)$. Then, a diffusion process is applied to $z$, and the reconstructed $x$ is obtained by using a decoder $\mathcal{D}(\cdot)$. In our setup, due to the small dimensions ($H$ and $W$) of the traffic ``image", we ignore $\mathcal{E}(\cdot)$ and $\mathcal{D}(\cdot)$, and take $x$ as input directly. We retained the denoising network of the LDM considering the effectiveness of the combination of ResNet and cross-attention employed by U-Net in the LDM. In LDM, the features from the conditional encoder are applied through the cross-attention mechanism, which is formulated as
\begin{equation}
    \mathcal{A}=\operatorname{Softmax}\left(\frac{Q K^{T}}{\sqrt{d}}\right),
\end{equation}
where $Q=W_{Q}^{(i)} \varphi_{i}\left(x_{t}\right)$, $K=W_{K}^{(i)} \mathcal{E}_{\mathcal{T}}\left(y\right)$. $\varphi_{i}\left(x_{t}\right)$ is an intermediate representation of U-Net. $W_{Q}^{(i)}$ and $W_{K}^{(i)}$ are learnable parameters. $\mathcal{E}_{\mathcal{T}}\left(\cdot\right)$ is the conditional encoder, where we use BERT \cite{43} to extract the text embedding.

\subsubsection{Graph convolutional network}
Diversity and consistency are two opposing goals when sampling conditional generative models. The image generation tasks, such as text-to-image generation, often prioritize diverse results. However, the TTG task places a higher emphasis on generative consistency. Specifically, the TTG task aims to produce consistent traffic situations when generating from the same textual input. To achieve this consistency, we integrate GCN into the diffusion model to provide stronger guidance by introducing spatial information of the road network. In the constructed dataset, we provide a neighborhood matrix $A$ to represent the spatial association of all roads. Given the adjacency matrix $A$ and data $x_{t}$, a two-layer GCN can be represented as
\begin{equation}
    f\left(x_{t}, A\right)=\sigma\left(\hat{A} \operatorname{ReLU}\left(\hat{A} x_{t} W_{0}\right) W_{1}\right),
\end{equation}
where $\hat{A}=\widetilde{D}^{-\frac{1}{2}} \widetilde{A} \widetilde{D}^{-\frac{1}{2}}$, $\widetilde{A}=A+I_{N}$. $\widetilde{D}$ is the degree matrix of $\widetilde{A}$, and $I_{N}$ is the unit matrix. $W_{0}$ and $W_{1}$ are the weight matrices. $\sigma\left(\cdot\right)$ is the activation function, and the sigmoid is used here. We further reformulate the optimization objective of ChatTraffic to 
\begin{equation}
    L_{CT}=\mathbb{E}_{x, y, \epsilon, t}\left[\left\|\epsilon-\epsilon_{\theta}\left(f\left(x_{t}, A\right), t, \mathcal{E}_{\mathcal{T}}(y)\right)\right\|_{2}^{2}\right],
\end{equation}
where $y$ denotes text, and $\epsilon_{\theta}\left(f\left(x_{t}, A\right), t, \mathcal{E}_{\mathcal{T}}(y)\right)$ is a series of denoising functions implemented via U-Net.

\subsubsection{Data visualization}
After completing the training of ChatTraffic, given a piece of text describing the traffic situation, we get the predicted traffic data $\hat{x} \in \mathbb{R}^{H \times W \times d}$. Each pixel point in $\hat{x}$ corresponds to a specific road, and we can visualize the traffic situation of these roads on a map, as shown in Figure \ref{fig1}. We use three different color bars to represent three different features, where speed and passing time are continuous data and congestion level is discrete data.

\begin{table*}[t]
\renewcommand{\arraystretch}{1.1}
\caption{Quantitative comparison of ChatTraffic with traditional traffic prediction methods on three specific junctures. Traditional methods predict data for the 20th, 40th, and 60th minutes in the future by inputting one hour of historical data. ChatTraffic directly inputs the text corresponding to the three junctures. The best performance is in bold.}
\centering
\resizebox{\textwidth}{!}
{\begin{tabular}{c|c|cccccc}
\toprule[1pt]
\multirow{2}{*}{Method} & \multirow{2}{*}{Input}                                                                                                                & \multicolumn{2}{c}{January 21, 2022, 18:20} & \multicolumn{2}{c}{January 21, 2022, 18:40} & \multicolumn{2}{c}{January 21, 2022, 19:00} \\ \cline{3-8} 
                        &                                                                                                                                       & \hspace{0.8em} MAE                  & RMSE                 & \hspace{0.8em} MAE                  & RMSE                 & \hspace{0.8em} MAE                  & RMSE                 \\ \midrule[1pt]
STGCN \cite{1}                   & \multirow{6}{*}{\begin{tabular}[c]{@{}c@{}}Historical data during \\ January 21, 2022, 17:00 -\\ January 21, 2022, 18:00\end{tabular}} & \hspace{1.0em}3.94                 & 6.04                 & \hspace{1em}4.37                 & 6.51                 &\hspace{0.8em} 4.69                 & 7.12                 \\
GWN \cite{2}                    &                                                                                                                                       & \hspace{1.0em}3.71                 & 5.68                 & \hspace{1em}3.96                 & 6.09                 &\hspace{0.8em} 4.11                 & 6.32                 \\
STSGCN \cite{44}                  &                                                                                                                                       & \hspace{1.0em}3.66                 & \textbf{5.55}                 & \hspace{1em}4.09                 & 6.08                 &\hspace{0.8em} 4.22                 & 6.41                 \\
ASTGCN \cite{50}                  &                                                                                                                                       & \hspace{1.0em}3.68                & 5.61                 & \hspace{1em}4.06                 & 6.17                 &\hspace{0.8em} 4.18                 & 6.39                 \\
MTGNN \cite{3}                  &                                                                                                                                       &\hspace{0.7em} \textbf{3.63}        & 5.57        & \hspace{1.0em}3.89                 & 5.95                 &\hspace{0.8em} 4.02                 & 6.13                 \\
STG-NCDE \cite{45}               &                                                                                                                                       &\hspace{0.7em} 3.65                 & 5.60                 & \hspace{1.0em}3.84                 & 5.89                 & \hspace{1.05em}4.09                 & 6.21                 \\ \midrule[0.5pt]
LDM \cite{13}            & \multirow{2}{*}{Text}                                                                                                                                  & \hspace{0.65em} 3.81                 & 5.84                 & \hspace{0.7em} 3.56        & 5.61        &\hspace{0.8em} 3.60        & 5.56   \\
ChatTraffic             &                                                                                                                                   & \hspace{0.65em} 3.73                 & 5.76                 & \hspace{0.7em} \textbf{3.38}        & \textbf{5.14}        &\hspace{0.8em} \textbf{3.44}        & \textbf{5.25}        \\ \bottomrule[1pt]
\end{tabular}}
  \label{table1}
\end{table*}

\section{Experiments and Analysis}
\label{section4}
\subsection{Experimental Setup}
\subsubsection{Dataset}
We construct a substantial dataset and carry out experiments on it since there is no publicly available dataset suitable for the TTG task. The dataset covers 1,260 roads within the fifth ring road area of Beijing, providing 22,320 text-traffic pairs. Each pair contains the traffic data $x$, where $x \in \mathbb{R}^{1260 \times 3}$, and a corresponding piece of text describing the state of the traffic system. We set 80\% of the entire dataset as the training set and the remaining 20\% as the testing set. The data collection interval is 4 minutes. The three dimensions of traffic data are the speed, the passing time of vehicles on each road, and the congestion level of each road. The text includes the time as well as the type and location of occurring abnormal events, which is structured as follows:
\begin{itemize}
\item[$\bullet$] March 21, 2022, 18:00. Road closure on south second ring road. A general traffic accident on s50 east fifth ring road. Road closure on tuanhe road. Construction and road closure on jingliang road. A general traffic accident on west fourth ring middle road. Road traffic control and a general traffic accident on s12 airport expressway. ......
\end{itemize}

\subsubsection{Compared methods}
To demonstrate the viability and effectiveness of our ChatTraffic, we choose five deep-learning methods for traditional traffic prediction to compare with our method. The five methods are STGCN \cite{1}, GWN \cite{2}, STSGCN \cite{44}, ASTGCN \cite{50}, MTGNN \cite{3} and STG-NCDE \cite{45}. It is worth noting that in the testing phase, the traditional traffic prediction methods rely on historical traffic data as input, whereas ChatTraffic utilizes text as input. 

\subsubsection{Evaluation metrics}
To evaluate the performance of all methods, we employ two standard metrics. Mean Absolute Error (MAE) and Root Mean Squared Error (RMSE) are used to quantify the prediction accuracy:
\begin{equation}
    M A E=\frac{1}{N} \sum_{i=1}^{N}\left|x_{i}-\hat{x}_{i}\right|,
\end{equation}
\begin{equation}
    R M S E=\sqrt{\frac{1}{N} \sum_{i=1}^{N}\left|x_{i}-\hat{x}_{i}\right|^{2}},
\end{equation}
where MAE calculates the average distance between the generated value $\hat{x}_{i}$ and the real value $x_{i}$, providing an assessment of the global quality of the generated traffic situation. RMSE evaluates the local quality of the generated traffic situation since it is more sensitive to outliers and noise. $N$ represents the number of predicted moments.

\subsubsection{Implementation details}
For traditional traffic prediction methods, we use the default parameters for training. For ChatTraffic, we train it with $T$ = 1000 noising steps and a linear noise schedule. The linear noise schedule starts at 0.00085 and ends at 0.012. The base learning rate is set to $10^{-5}$ while the batch size is set to 4. All the experiments are conducted on a single Nvidia GeForce 4090 ($\sim$ 24GB).

\begin{figure*}[h!]
  \centering
   \includegraphics[width=\textwidth]{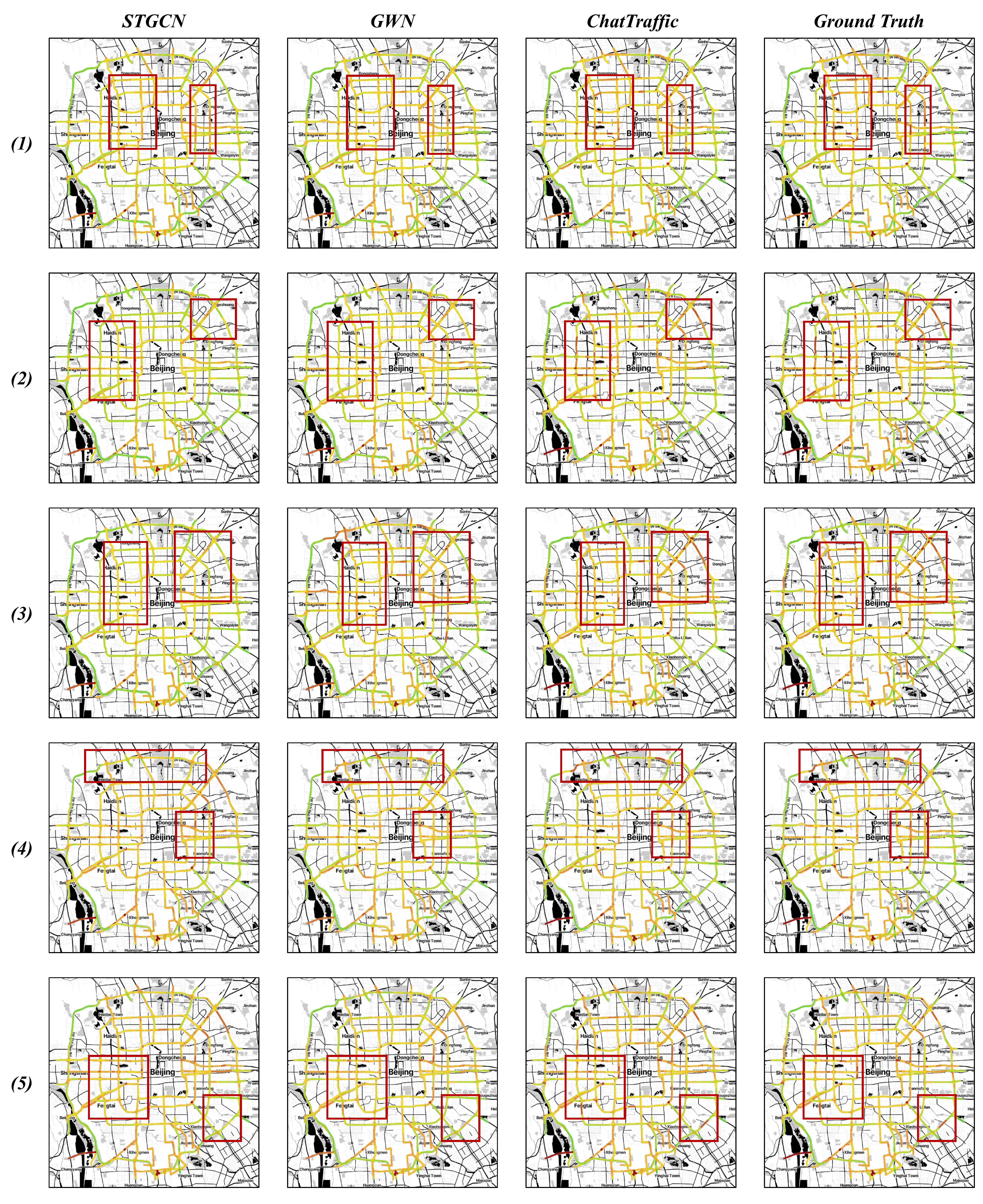}
   \caption{Qualitative comparison of ChatTraffic with two traditional traffic prediction methods on five specific junctures. The first to fifth rows represent five specific junctures. Best viewed in red boxes.}
   \label{fig5}
\end{figure*}

\begin{table*}[h!]
\renewcommand{\arraystretch}{1.3}
  \centering
\caption{Ablation results on GCN. ChatTraffic generates the most accurate predictions when the number of layers of the GCN is set to 2. Number of samples indicates how many samples to produce for the given prompt. As the number of samples generated increases, the time taken to generate the samples also increases. The best performance is in bold.}
  % \resizebox{\textwidth}{!}
{\begin{tabular}{c|c|c|cccccc}
\toprule[1pt]
\multirow{2}{*}{Layers of GCN} & \multirow{2}{*}{Numbers of samples} & \multirow{2}{*}{Time consumption (sec)} & \multicolumn{2}{c}{Congestion level} & \multicolumn{2}{c}{Speed (km/h)} & \multicolumn{2}{c}{Passing time (sec)} \\ \cline{4-9} 
                               &                                     &                                  & MAE               & RMSE             & MAE             & RMSE           & MAE              & RMSE              \\ \midrule[1pt]
\multirow{3}{*}{0}             & 1                                   & 9.76                              & 0.07              & 0.26             & 4.42            & 6.47           & 105.73           & 216.21            \\
                               & 5                                   & 45.14                              & 0.06              & 0.21             & 3.46            & 5.08           & 83.29            & 169.93            \\
                               & 10                                  & 92.93                              & 0.05              & 0.20             & 3.31            & 4.92           & 79.83            & 164.12            \\ \midrule[1pt]
\multirow{3}{*}{1}             & 1                                   & 9.81                              & 0.06              & 0.24             & 4.31            & 6.35           & 100.21           & 209.52            \\
                               & 5                                   & 44.97                              & 0.05              & 0.20             & 3.35            & 5.01           & 80.74            & 166.97            \\
                               & 10                                  & 93.14                              & 0.05              & 0.19             & 3.26            & 4.86           & 77.19            & 161.26            \\ \midrule[1pt]
\multirow{3}{*}{\textbf{2}}    & 1                                   & 9.79                              & 0.06              & 0.24             & 4.23            & 6.29           & 98.98            & 204.54            \\
                               & 5                                   & 46.62                              & 0.05              & 0.19             & 3.35            & 4.98           & 79.01            & 164.59            \\
                               & \textbf{10}                         & 93.68                              & \textbf{0.05}     & \textbf{0.18}    & \textbf{3.21}   & \textbf{4.78}  & \textbf{75.89}   & \textbf{158.92}   \\ \midrule[1pt]
\multirow{3}{*}{3}             & 1                                   & 10.03                              & 0.06              & 0.26             & 4.35            & 6.28           & 102.75           & 213.78            \\
                               & 5                                   & 47.31                              & 0.05              & 0.19             & 3.38            & 4.98           & 82.31            & 168.32            \\
                               & 10                                  & 94.25                              & 0.05              & 0.19             & 3.29            & 4.86           & 78.23            & 163.51            \\ \bottomrule[1pt]
\end{tabular}}
  \label{table3}
\end{table*}

\begin{table}[h!]
\renewcommand{\arraystretch}{1.3}
  \centering
\caption{Quantitative comparison of ChatTraffic with traditional traffic prediction methods on the test set.}
  \resizebox{\columnwidth}{!}
{\begin{threeparttable}
\begin{tabular}{c|cccccc}
\toprule[1pt]
\multirow{2}{*}{Method} & \multicolumn{2}{c}{H=5\tnote{1}} & \multicolumn{2}{c}{H=10\tnote{2}} & \multicolumn{2}{c}{H=15\tnote{3}} \\ \cline{2-7} 
                        & MAE        & RMSE       & MAE         & RMSE       & MAE         & RMSE       \\ \midrule[1pt]
STGCN \cite{1}                  & 3.86       & 5.85       & 4.40        & 6.55       & 4.45        & 6.83       \\
GWN \cite{2}                    & 3.56       & 5.51       & 3.78        & 5.98       & 3.95        & 6.23       \\
STSGCN \cite{44}                 & 3.73       & 5.70       & 4.07        & 6.21       & 4.25        & 6.65       \\
ASTGCN \cite{50}                 & 3.80       & 5.84       & 4.06        & 6.34       & 4.10        & 6.43       \\
MTGNN \cite{3}                  & 3.49       & 5.39       & 3.85        & 5.94       & 4.28        & 6.14       \\
STG-NCDE \cite{45}               & 3.40       & 5.17       & 3.88        & 6.23       & 4.03        & 6.51       \\ \midrule[1pt]
\multirow{2}{*}{Method} & \multicolumn{6}{c}{Test set\tnote{4}}                                                  \\ \cline{2-7} 
                        & \multicolumn{3}{c}{MAE}               & \multicolumn{3}{c}{RMSE}              \\ \midrule[1pt]
LDM \cite{13}             & \multicolumn{3}{c}{3.31}     & \multicolumn{3}{c}{4.92} \\
ChatTraffic             & \multicolumn{3}{c}{\textbf{3.21}}     & \multicolumn{3}{c}{\textbf{4.78}}     \\ \bottomrule[1pt]
\end{tabular}
\begin{tablenotes}    
        \footnotesize               
        \item[1,2,3] For traditional prediction methods, we take 15 data as input, for a total time span of 1 hour. We then make predictions for the 5th data (20 minutes), the 10th data (40 minutes), and the 15th data (60 minutes).          
        \item[4] ChatTraffic accepts the text corresponding to each data in the test set as input and directly generates predictions for each specific moment.        
      \end{tablenotes}
\end{threeparttable}}
  \label{table2}
\end{table}

\subsection{Quantitative and Qualitative Comparisons}
We quantitatively compare our proposed ChatTraffic with several state-of-the-art traffic prediction methods. The experimental results are listed in Table \ref{table1}. For the traditional traffic prediction method, we choose to input 15 consecutive data from the test set, covering a 1-hour duration, and make predictions for the subsequent 5, 10, and 15 data. 

With ChatTraffic, we directly input the text corresponding to the three specific junctures that require predictions. Table \ref{table1} illustrates that the performance of traffic prediction methods deteriorates as time increases. In contrast, ChatTraffic consistently maintains a low and stable prediction error across all three junctures. We then quantitatively compare ChatTraffic with traffic prediction methods on the entire test set, as illustrated in Table \ref{table2}. Similarly, we employ 15 consecutive data as input for the traditional methods and move through the entire test set using the sliding window approach. Table \ref{table1} and Table \ref{table2} demonstrate that ChatTraffic not only matches the performance of state-of-the-art traffic prediction methods in short-term predictions but also maintains such performance in long-term predictions. This evidence confirms that the point-to-point generation strategy employed by ChatTraffic remains unaffected by the duration of the prediction period. In other words, ChatTraffic effectively alleviates the challenge of limited performance in long-time prediction of current traffic prediction methods. Consequently, ChatTraffic holds the capability to generate future traffic situations, especially in scenarios influenced by abnormal events.

To intuitively observe the ability of ChatTraffic to perceive abnormal road events, we also conduct a qualitative analysis. In Figure \ref{fig5}, we present a comprehensive visual comparison of ChatTraffic with two representative methods. The first to fifth rows represent five specific junctures. From the red boxes, it can be observed that outputs of ChatTraffic closely align with the ground truth, demonstrating its capacity to reflect the influence of abnormal events on roads. On the contrary, the two traditional methods seldom succeed in predicting anomalies, such as individual road congestion. These visualizations demonstrate that ChatTraffic is more sensitive to traffic anomalies compared to traditional traffic prediction methods, resulting in more accurate predictions of traffic situations.

\begin{figure*}[h!]
  \centering
   \includegraphics[width=\textwidth]{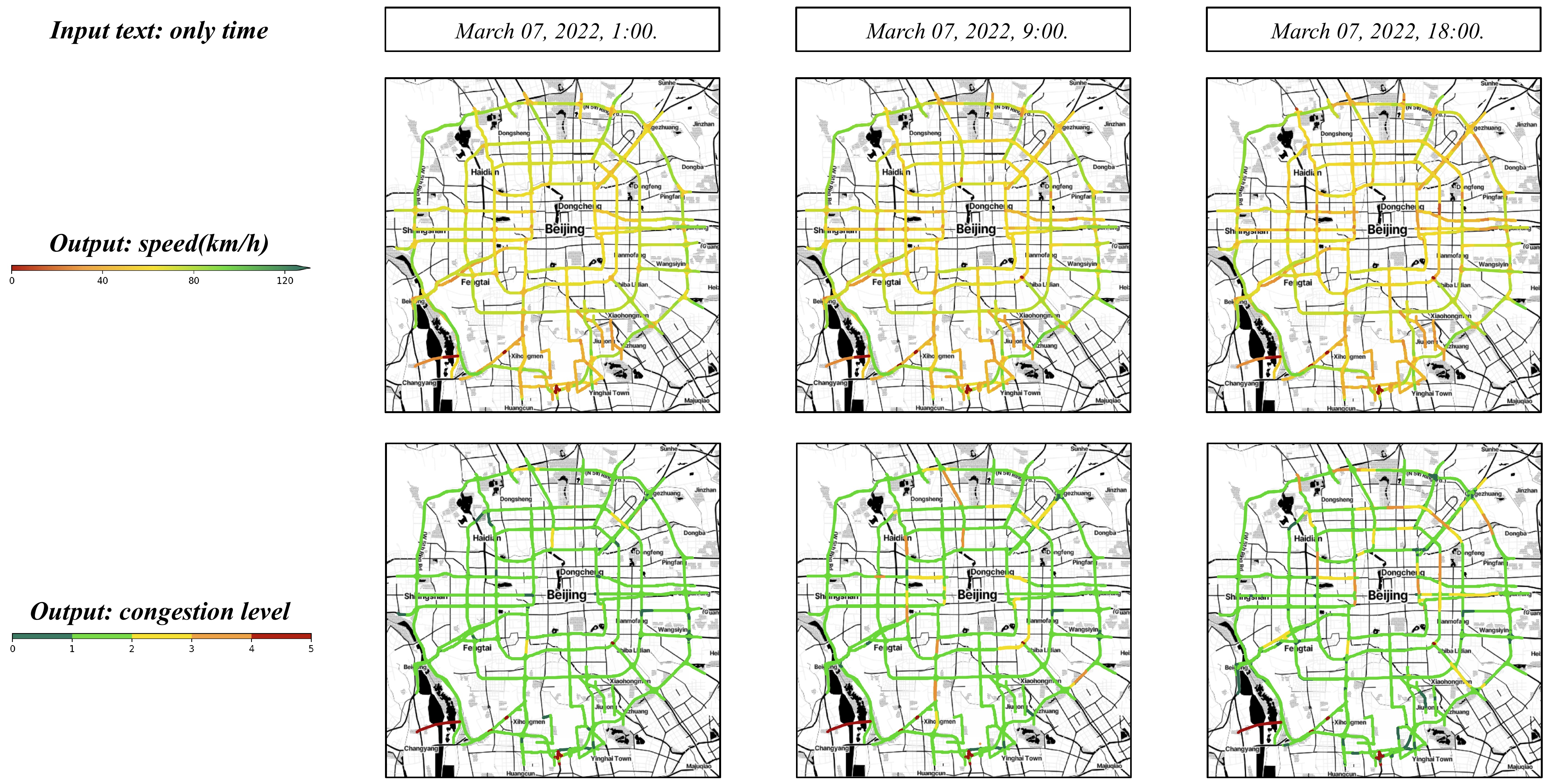}
   \caption{Output visualization. ChatTraffic can predict routine traffic patterns, i.e., the morning and evening peaks, when the input text contains only time.}
   \label{fig6}
\end{figure*}

\subsection{Ablation Study}
\subsubsection{Effectiveness of GCN}
In order to understand the effectiveness of GCN, we conduct an ablation experiment on the number of layers of GCN in ChatTraffic. Results are presented in Table \ref{table3}. Incorporating Graph Convolutional Networks (GCN) into ChatTraffic has been proven to enhance its generation accuracy, with the minimum generation error observed when using two layers of GCN. It's essential to highlight that as the number of GCN layers continues to increase, the error begins to rise instead. This phenomenon can be attributed to the road network's adjacency matrix having a notably sparse structure. When the sparsity of this adjacency matrix is low, there is a risk of over-smoothing, rendering the utilization of multi-layer GCN less effective. Besides, the time consumption of ChatTraffic to generate a single prediction sample is approximately ten seconds. When five samples are generated and averaged based on a test prompt, the MAE and RMSE decrease substantially compared to when only one sample is generated. However, while generating ten samples and averaging them further reduces the MAE and RMSE, the improvement is not markedly significant. Therefore, generating five samples is the preferred approach when using ChatTraffic for an optimal balance between performance and time efficiency.

\begin{figure*}[t!]
  \centering
   \includegraphics[width=\textwidth]{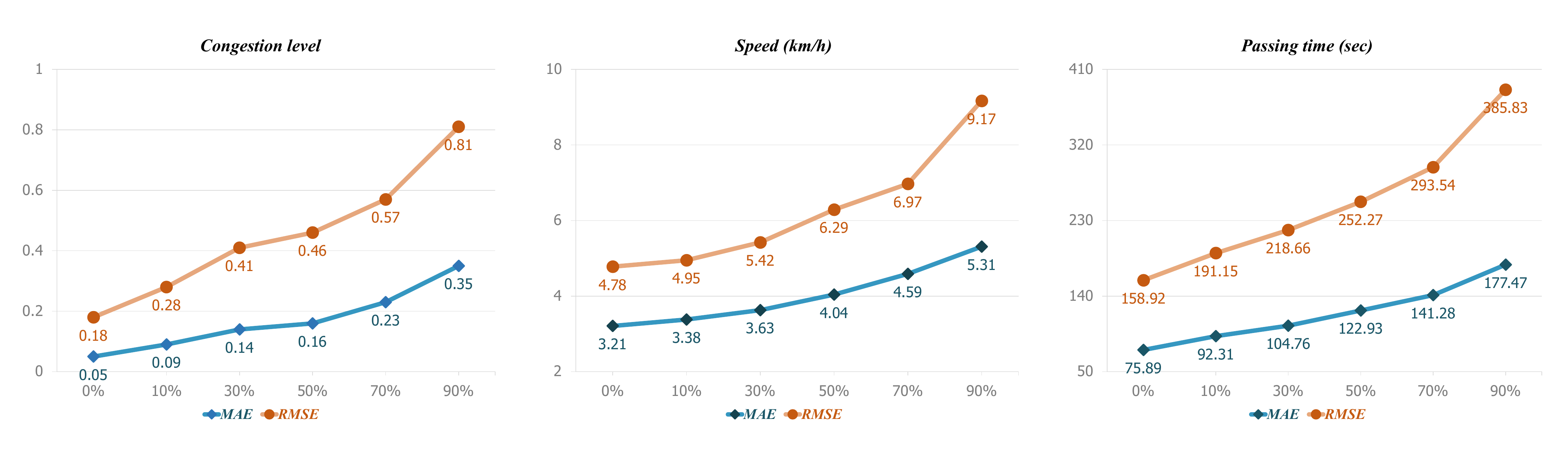}
   \caption{Ablation results on the missing rate of input events prompt. The missing rate is set to 10\%, 30\%, 50\%, 70\%, and 90\%. As the missing rate of input event increases, there is a corresponding rise in the MAE and RMSE of the three generated traffic data.}
   \label{fig7}
\end{figure*}

\subsubsection{Input partial events}
To verify how the quantity of input events impacts the traffic situations generated by ChatTraffic, we use incomplete event prompts from the test set as inputs in this ablation study. We randomly erase events from each complete event prompt in the test set with the missing rates of 10\%, 30\%, 50\%, 70\%, and 90\%. The results are illustrated in Figure \ref{fig7}, showing that as the missing rate of input event increases, there is a corresponding rise in the MAE and RMSE of the three generated traffic data. This phenomenon proves that the traffic situations generated by ChatTraffic have a strong correlation with the input event data. It is worth noting that the increases in MAE and RMSE are relatively stable and remain within a low range, without any drastic spikes. This observation suggests that ChatTraffic might have the ability to generate traffic situations that align with the cyclic patterns of the traffic system based solely on temporal factors, even without any specific event input. To further explore the hypothesis, we conduct experiments in the next ablation study.

\subsubsection{Input time only}
In this study, we aim to demonstrate that ChatTraffic can predict routine traffic patterns, such as the morning and evening peaks, similar to traditional traffic prediction methods. Therefore, we input text containing only time information to ChatTraffic. Figure \ref{fig6} illustrates the speed and road congestion level generated by ChatTraffic at three different junctures: 1:00, 9:00, and 18:00. Based on conventional knowledge, traffic congestion should be minimal at 1:00, while 9:00 and 18:00 correspond to the morning and evening rush hours, during which certain roads are expected to experience congestion. From the data represented in Table \ref{fig6},  it is evident that ChatTraffic generates results that align closely with this anticipated pattern. It's noteworthy that the speeds of most roads at these three junctures are relatively modest. This is primarily because of the presence of speed limits on urban roads. Therefore, it can be reasonably inferred that ChatTraffic can deduce the influence of time variations on the traffic system from text containing only time, even in the absence of historical data.

\section{Conclusion}
\label{section5}
In this paper, we explore a novel multimodal traffic prediction task called Text-to-Traffic Generation, which aims to generate traffic situations described by text. To that end, we propose ChatTraffic, the first diffusion-based text-to-traffic generative model, and construct a substantial text-traffic dataset. Different from traditional traffic prediction methods that only use historical data, combining the advancement of the diffusion model and the spatial perception offered by the GCN, ChatTraffic is armed with realistic and accurate traffic generation capability. We demonstrate the superiority of ChatTraffic over traditional prediction methods by conducting comparison experiments on the constructed dataset and showing the visualizations. We anticipate a growing interest from researchers in the TTG domain due to its significant practical applications, notably in enhancing the efficiency of urban transportation system management. In the future, we will further explore improving the accuracy of traffic situation generation through multimodal learning.

{
\bibliographystyle{IEEEtran}
\bibliography{main}
}

\begin{IEEEbiography}[{\includegraphics[width=1in,height=1.25in,clip,keepaspectratio]{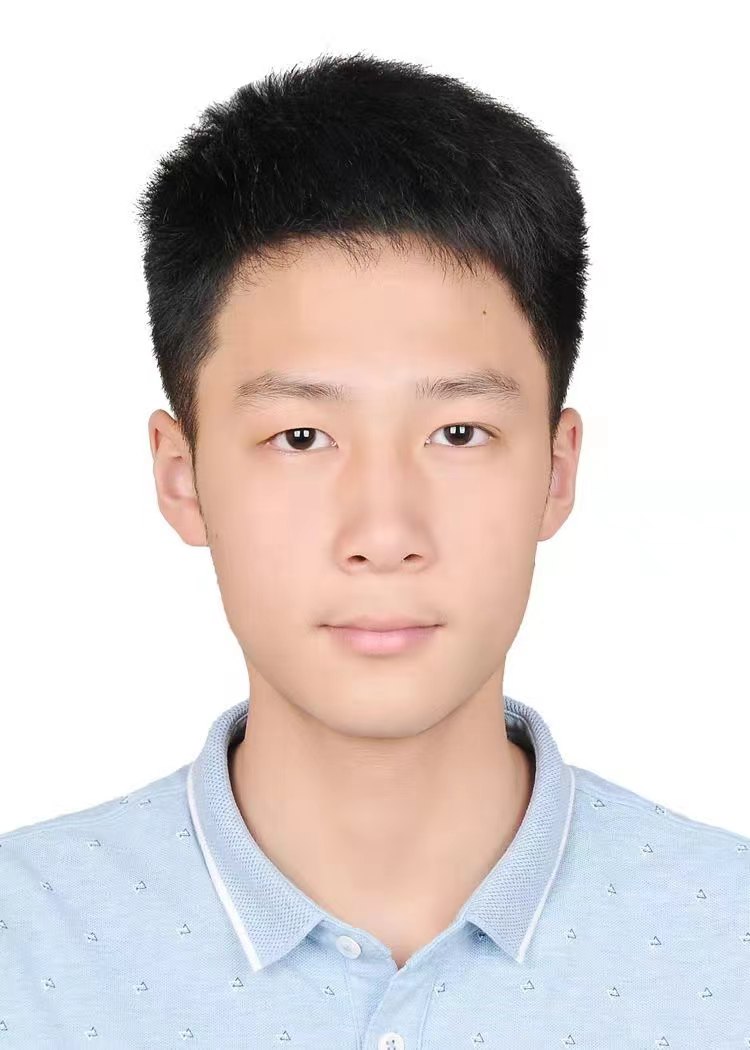}}]{Chengyang Zhang} received his bachelor degrees from Beijing Information Science and Technology University. He is currently a graduate student at the Department of Informatics, Beijing University of Technology. His research interests include computer vision, traffic prediction and biomedical image analysis.
\end{IEEEbiography}

\begin{IEEEbiography}[{\includegraphics[width=1in,height=1.25in,clip,keepaspectratio]{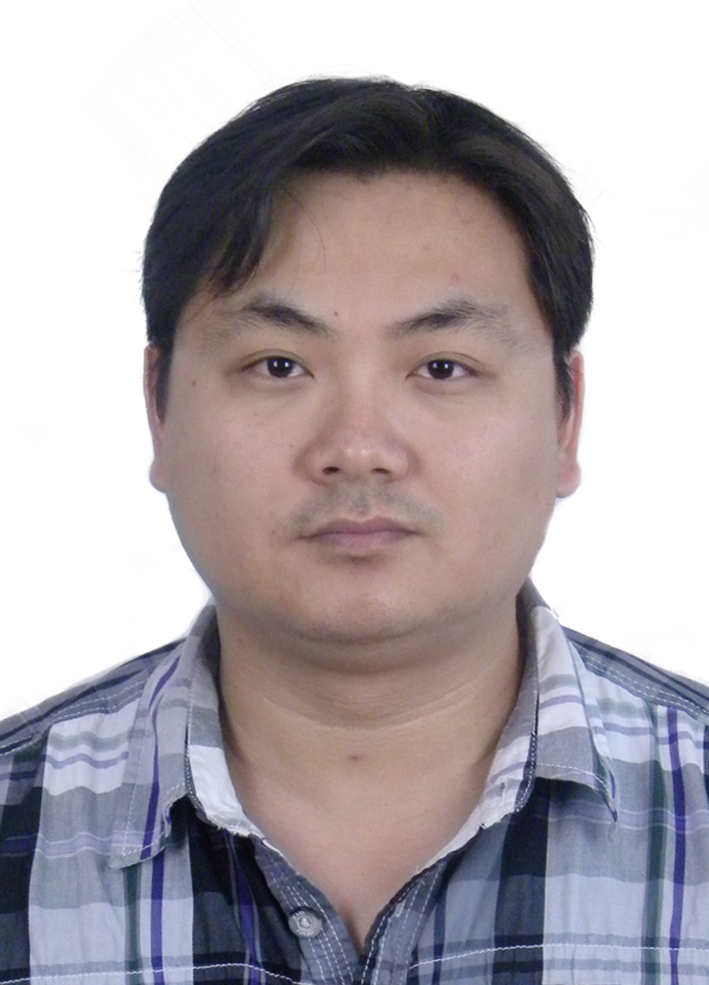}}]{Yong Zhang} (Member, IEEE) received the Ph.D.degree in computer science from the Beijing University of Technology in 2010. He is currently an Associate Professor in computer science with the Beijing University of Technology. His research interests include intelligent transportation systems, big data analysis, visualization, and computer graphics.
\end{IEEEbiography}

\begin{IEEEbiography}[{\includegraphics[width=1in,height=1.25in,clip,keepaspectratio]{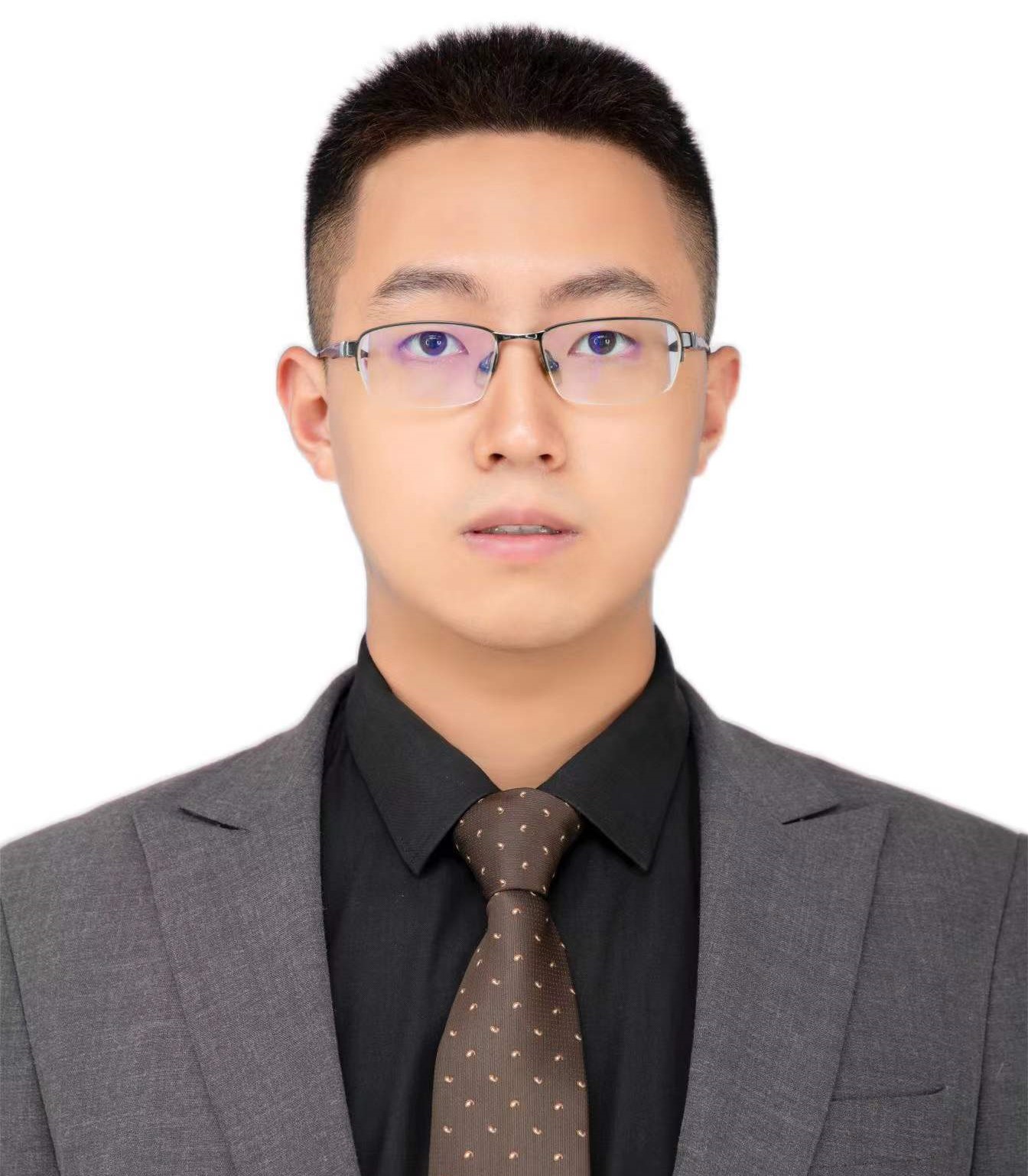}}]{Qitan Shao} received the bachelor’s degree in computer science and technology from the Beijing University of Technology. He is currently a graduate student at the Department of Informatics, Beijing University of Technology. His research interest is intelligent transportation systems.
\end{IEEEbiography}

\begin{IEEEbiography}[{\includegraphics[width=1in,height=1.25in,clip,keepaspectratio]{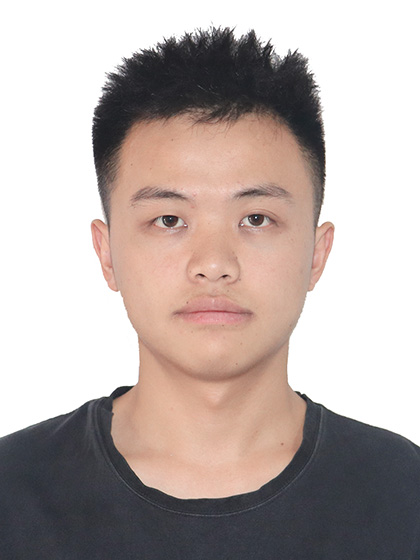}}]{Bo Li} received his bachelor degrees from Beijing Information Science and Technology University. He is currently a graduate student at the Department of Informatics, Beijing University of Technology. His research interests include computer vision and biomedical image analysis.
\end{IEEEbiography}

\begin{IEEEbiography}[{\includegraphics[width=1in,height=1.25in,clip,keepaspectratio]{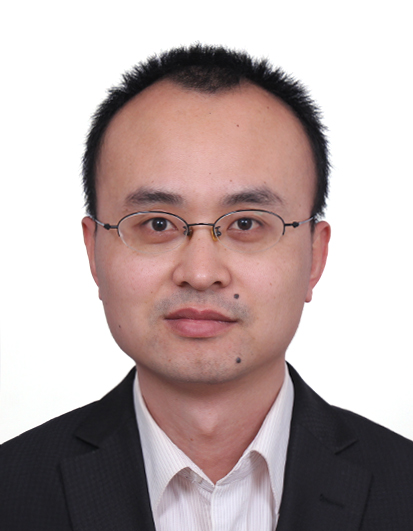}}]{Yisheng Lv} received the B.E. and M.E. degrees in transportation engineering from Harbin Institute of Technology, Harbin, China, in 2005 and 2007, respectively, and the Ph.D. degree in control theory and control engineering from Chinese Academy of Sciences, Beijing, China, in 2010. He is an Assistant Professor with State Key Laboratory of Management and Control for Complex Systems, Institute of Automation, Chinese Academy of Sciences. His research interests include traffic data analysis, dynamic traffic modeling, and parallel traffic management and control systems.
\end{IEEEbiography}

\begin{IEEEbiography}[{\includegraphics[width=1in,height=1.25in,clip,keepaspectratio]{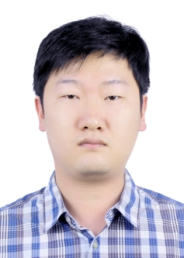}}]{Xinglin Piao} received the Ph.D. degree from the Beijing University of Technology, Beijing, China, in 2017. He is currently a lecturer in the Faculty of Information Technology at Beijing University of Technology. His research interests include intelligent traffic, pattern recognition, and multimedia technology.
\end{IEEEbiography}

\vspace{-16cm}

\begin{IEEEbiography}[{\includegraphics[width=1in,height=1.25in,clip,keepaspectratio]{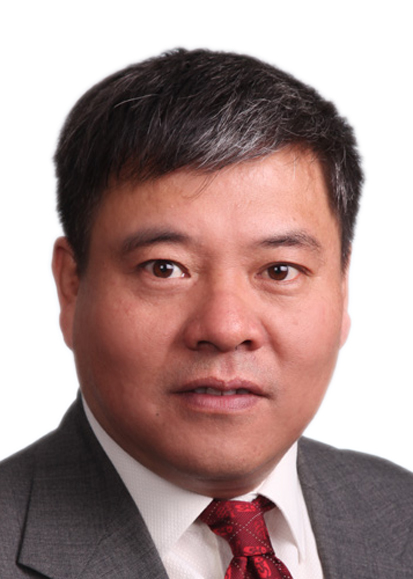}}]{Baocai Yin} (Member, IEEE) received the B.S., M.S., and Ph.D. degrees in computational mathematics from the Dalian University of Technology, Dalian, China, in 1985, 1988, and 1993, respectively. He is currently a Professor with the Beijing Key
Laboratory of Multimedia and Intelligent Software Technology, Faculty of Information Technology, Beijing University of Technology. His research interests include multimedia, image processing, computer vision, and pattern recognition.
\end{IEEEbiography}

\end{document}